\documentclass[sigplan,10pt,nonacm]{acmart}
\setcopyright{none}
\renewcommand\footnotetextcopyrightpermission[1]{}
\acmConference{}{}{}
\acmDOI{}
\acmISBN{}
\usepackage{amsmath}
\usepackage{booktabs}
\usepackage{graphicx}
\usepackage{tabularx}
\usepackage{microtype}
\graphicspath{{figures/}{./}}

\newcommand{\sd}{\operatorname{SD}}

\newif\ifincludeappendix
\includeappendixtrue % Set false when the venue requires a main-text-only PDF.

\begin{document}
\title[Mi-Ripple: Restoring Images Degraded by Iterative AI Editing]{Mi-Ripple: Restoring Images Degraded by Iterative AI Editing}
\author{Jiayin Chen\textsuperscript{2,1,3,4}, Yicheng Xu\textsuperscript{1,5}, Muting Wang\textsuperscript{1,6}}
\affiliation{\institution{\textsuperscript{1} Miyang Technology (Shanghai) Co., Ltd., Shanghai, China; \textsuperscript{2} Key Laboratory of System Software (Chinese Academy of Sciences), Beijing, China; \textsuperscript{3} Institute of Software, Chinese Academy of Sciences, Beijing, China; \textsuperscript{4} University of Chinese Academy of Sciences, Beijing, China; \textsuperscript{5} Shanghai Jiao Tong University, Shanghai, China; \textsuperscript{6} Tianjin University, Tianjin, China}\country{China}}

\begin{abstract}
Iterative reference-conditioned image editing can introduce grid-like and granular textures, commonly described as digital ripple.
We present Mi-Ripple, a diagnosis-guided restoration workflow that suppresses this digital ripple while protecting image structure. 
Mi-Ripple separates periodic lattice artifacts from content-entangled granular texture, then combines selective spectral notching, 
structure-aware smoothing, and cleaned-reference regeneration. 
This separation enables low-distortion filtering when artifacts are spectrally isolated and visual reconstruction when filtering would erase legitimate detail. 
Across fourteen notch-only executions, whole-image residual standard deviation is 0.08--0.44 in CIELAB lightness units. 
In a paired regeneration example, reference cleaning reduces output debris density by 45\%. 
Mi-Ripple links measurable artifact reduction to visibly cleaner generated images, rather than optimizing a spectral score alone.
\end{abstract}
\begin{teaserfigure}
  \centering
  \includegraphics[width=\textwidth]{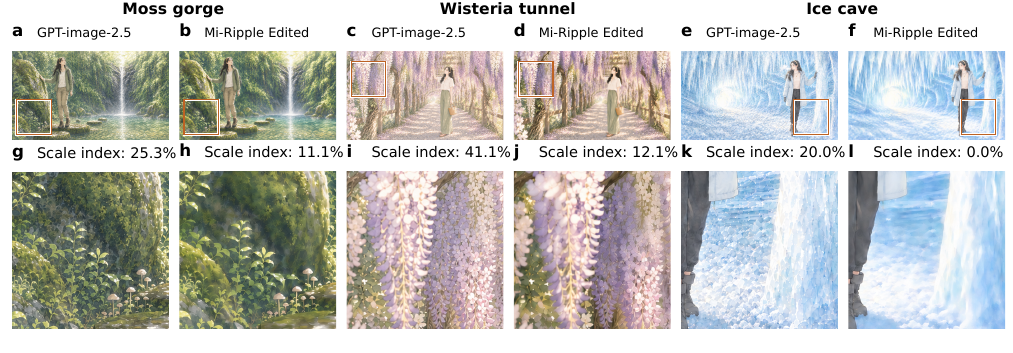}
  \caption{Restoring GPT-image-2.5 gen4 outputs with reference cleaning, one regeneration per scene, and final notching. Normalized scale-tile percentages decrease from 25.3\% to 11.1\% in moss gorge (H), 41.1\% to 12.1\% in wisteria tunnel (W), and 20.0\% to 0.0\% in ice cave (K).}
  \Description{Full-frame and relative-location crop pairs compare GPT-image-2.5 moss gorge, wisteria tunnel, and ice-cave endpoints with complete restoration outputs. Scale coverage falls but remains structured in moss and wisteria, and reaches zero in the ice cave. Regeneration changes scene details.}
  \label{fig:teaser}
\end{teaserfigure}
\maketitle

\section{Introduction}
Iterative reference-conditioned image editing uses an existing image and a text instruction to preserve visual identity while modifying a scene. 
Reusing each output as the next reference makes incremental editing convenient, but can also propagate unwanted texture. 
At native resolution, affected images exhibit grids, honeycomb-like patterns, or granular surfaces that are inconspicuous in thumbnails. 
We use \emph{digital ripple} as an umbrella term for these structured artifacts.

Prior work has shown that generated images can exhibit systematic frequency-domain discrepancies, with several studies linking these patterns to upsampling, aliasing, and related sampling effects~\cite{durall,dzanic,frank,odena,karras}. This line of research has mainly focused on detecting synthetic images or explaining artifact formation at the architectural level. Repeated reference-conditioned editing raises a more practical restoration problem: whether a visible artifact is spectrally separable and can be removed without harming legitimate detail, or whether it is entangled with scene content and therefore requires intervention in the generation workflow. Studies of repeated image replication, including Banana100, further show that recursive degradation challenges no-reference quality assessment~\cite{banana100}. These findings motivate a treatment-oriented analysis that distinguishes removable spectral contamination from content-entangled degradation.

Our goal is to recover clean surfaces and coherent detail without erasing legitimate texture (Figure~\ref{fig:teaser}).
The contributions are as follows:

\begin{itemize}
  \item \textbf{Restoration beyond denoising.} We combine isolated-peak notching, structure-aware suppression, and cleaned-reference regeneration to repair both spectral contamination and degraded texture.
  \item \textbf{Diagnosis-driven treatment.} Spectral and spatial probes distinguish lattice from granular artifacts and select filtering, regeneration, or review.
  \item \textbf{Visual and measured evidence.} Restoration pairs and six portrait cases show the workflow's benefits; aligned residuals independently check filtering damage.
\end{itemize}

\section{Related Work}
Checkerboard artifacts can arise from uneven overlap in deconvolution~\cite{odena}. Subsequent studies use spectral discrepancies to detect generated images~\cite{durall,dzanic,frank,corvi}, while alias-free synthesis addresses sampling-related artifacts at the architectural level~\cite{karras}. We build on frequency analysis, but use it to localize a treatable component rather than classify an entire image as synthetic.

Self-consuming training can degrade generative distributions~\cite{mad,collapse}. Here, model parameters remain fixed and recursion occurs through the reference image at inference time. Banana100 is the closest empirical setting~\cite{banana100,banana100data}. Our analysis of its supplementary sequences separates persistent periodicity from changes in artifact strength and scene content.

Frequency-domain notch filtering is a classical treatment for periodic noise~\cite{gonzalez}. Our implementation selects isolated components relative to a two-dimensional local baseline and verifies spatial distortion after filtering. Watermarking can also introduce structured signals~\cite{treering,stablesig}. Output spectra alone therefore cannot establish whether a particular lattice originates in synthesis, resizing, or watermarking. Our restoration policy depends on the observed signal, not on resolving that attribution.

\section{Diagnosis-Guided Restoration}
\label{sec:mitigation}
Our workflow has three stages: diagnose the artifact, filter or regenerate from a cleaned reference, and verify the resulting candidate. It separates \emph{deliverable-grade} filtering, which preserves pixel alignment, from \emph{reference-grade} cleaning, which prepares an input for regeneration. This distinction permits stronger reference preparation without presenting it as a finished image.

\subsection{Diagnose Separable and Content-Entangled Artifacts}
\begin{figure}[t]
  \centering
  \includegraphics[width=\linewidth]{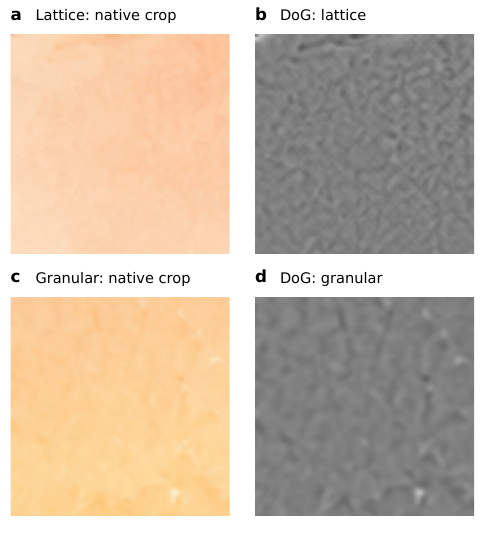}
  \caption{Native crops and difference-of-Gaussians (DoG) views of a lattice-positive candidate and a granular region.}
  \Description{Two native crops and their amplified difference-of-Gaussians views illustrate fine texture in a lattice-positive candidate and a granular region. Filter widths and display gains are labeled.}
  \label{fig:forms}
\end{figure}

Complementary spectral and spatial probes determine which restoration route is appropriate (Figure~\ref{fig:forms}). All lightness measurements use CIELAB $L^*$ on a 0--100 scale. For a lightness patch $P$ and Hann window $w$, define
\begin{equation}
 S=\log\!\left(1+\left|\mathcal{F}[(P-\bar P)w]\right|\right),
 \qquad A(u,v)=S(u,v)-B(r),
 \label{eq:anomaly}
\end{equation}
where $B(r)$ is the median log-amplitude at radius $r$. We summarize $A$ at radii $r\geq8$ frequency bins. The baseline survey uses the median peak across 384-pixel flat patches per image. Autocorrelation of high-passed lightness supplies a separate period estimate; a secondary peak near the photographic control level, approximately 0.05, is not treated as reliable.

The whole-image lattice probe instead uses a $21\times21$ local median of the log-amplitude spectrum. It retains connected components with excess above 1.2, outside radius 24, and support at most 80 frequency bins. Diagnosis requires at least two retained components and a largest excess of at least 2.5. This lattice score differs from $\max A$ in Equation~\ref{eq:anomaly}; their thresholds are not interchangeable.

A flat-window probe tests 96-pixel windows for band-pass strength, excess kurtosis, blob coverage, and isotropy. A complementary whole-frame \emph{scale index} reports the percentage of qualifying 128-pixel tiles at stride 64. Qualifying tiles combine sufficient blob coverage with low blob-area coefficient of variation, high circularity, and low anisotropy. The whole-frame probe covers dense foliage even when no flat window qualifies. Appendix~\ref{app:measurement} supplies the thresholds and grading rules.

The nominal 3--8-pixel band isolates diagnostic texture but also imposes a size preference. We therefore check periodicity separately, without that band-pass. Granular regions have weak autocorrelation maxima at inconsistent displacements, whereas the lattice produces repeatable period vectors. Directional rendering changes, such as ribbon-like hair, are treated separately from both forms.

\subsection{Filter or Regenerate from a Cleaned Reference}
For a lattice, isolated-peak notching uses the diagnostic probe's component-selection criterion. Let $K$ indicate retained components and let $E$ be their positive excess above the local baseline. With Gaussian feathering $G_{1.5}$, the filtered spectrum is
\begin{equation}
 \widehat F=F\exp[-G_{1.5}(K E)].
 \label{eq:notch}
\end{equation}
The operation preserves phase and normally modifies lightness alone. Reflection padding reduces boundary effects. Compact-peak selection avoids the extended spectral ridges associated with directional image content. The comparison method, radial-baseline soft clipping, instead suppresses sufficiently large deviations from the radial median.

Granular texture has no isolated peak for notching to remove. In unstructured regions, a strict mask based on edge strength, orientation coherence, and texture density permits local band reduction while protecting structure. Where artifacts overlap legitimate foliage or material texture, the pipeline requests human review rather than increasing filtering strength.

When content reconstruction is acceptable, reference-grade cleaning permits broader suppression before regeneration, with explicit face protection. The regenerated output is diagnosed again, and newly introduced isolated lattice peaks can be notched. This route treats regeneration as a new candidate: it may invent detail and requires visual inspection rather than aligned pixel comparison.

\subsection{Verify Distortion and Record the Decision}
Acceptance uses image differences rather than the anomaly score optimized by the filter. Structural windows must have residual standard deviation (SD) at most 0.6 lightness units and high-frequency retention of at least 90\%. Retention is the output-to-input ratio of $\sd(L-G_1L)$, where $G_1$ is Gaussian smoothing with standard deviation one pixel. Flat-window band-pass SD must not increase by more than 0.02, and whole-image residual SD must not exceed 1.0. These empirical checks measure filtering damage; human review determines whether to adopt the candidate.

Rule-based routing is the default implementation. An optional language-model layer selects among permitted actions without changing numerical thresholds. Execution records retain observations, allowed actions, decisions, and verification results. Regeneration requires explicit permission and a bounded retry count; Appendix~\ref{app:pipeline} distinguishes the implemented orchestration from the separately executed restoration examples.

\section{Experimental Setup}
\label{sec:methods}
\subsection{Data and Editing Protocol}
The study uses two commercial editing channels without access to weights, seeds, or sampling parameters. 
Channel A is a desktop integration advertised as OpenAI Image 2. 
Channel B is an OpenAI-compatible gateway including routes identified as gpt-image-2 and gpt-image-2.5. 
We treat these as sampled access conditions rather than independently replicated models or a vendor ranking.

Our experiments comprise five five-generation same-scene chains on Channel B, two five-generation scene-change chains on Channel A, prompt comparisons on two foliage-rich scenes, and an eight-scene comparison of the gpt-image-2 and gpt-image-2.5 routes. Each chain contains an initial output, gen0, and four edits, gen1--gen4. Same-scene chains reuse the previous output with an unchanged prompt. Additional material includes four photographs, five web references, fifteen generated images for a spectral survey, four fixed-condition repeats, and six watercolour portrait originals for the pipeline case series.

For external evidence, we analyze the Banana100 \texttt{more\_models} subset~\cite{banana100data}: one starting photograph and 110 edited outputs from seven model families. Its eleven ten-step sequences include different-chat and same-chat variants.

\subsection{Measurement Conventions}
Actual image dimensions are read from file headers. Periodicity is measured at native resolution. Cross-resolution comparisons use Lanczos downsampling, with native readings retained separately. The eight-scene comparison uses the same measurement code and normalizes the long edge to 1280 pixels. Each scene has one chain per route; successive generations are not independent replicates. Normalization does not eliminate differences in the outputs' native-canvas sampling histories.

\section{Restoration Evaluation}
\label{sec:restoration-evaluation}
\subsection{Visual Restoration and Reference Cleaning}
\begin{figure}[t]
\centering
\includegraphics[width=\linewidth]{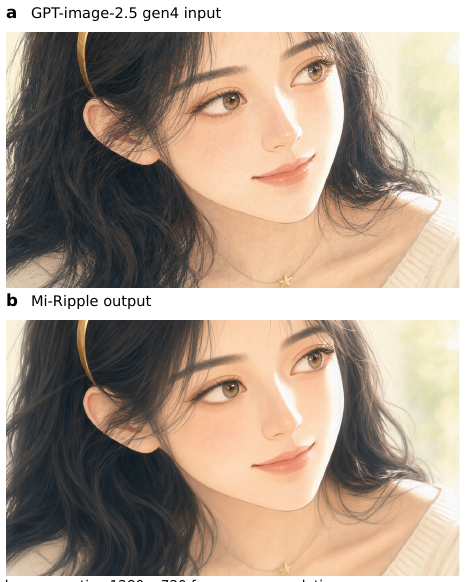}
\caption{GPT-image-2.5 gen4 face close-up and the Mi-Ripple output. The comparison illustrates the visual change after reference cleaning, regeneration and final notching.}
\Description{Two native 1280 by 720 face close-up images are stacked vertically. The top is a GPT-image-2.5 gen4 input and the bottom is the Mi-Ripple output after processing. No separate enlarged crop is shown.}
\label{fig:restoration}
\end{figure}

The restoration examples show how treatment selection addresses visible grain, fragmented detail, and contaminated reference texture (Figure~\ref{fig:teaser}). Reference cleaning followed by regeneration produces a reconstructed candidate even without access to a clean ancestor (Figure~\ref{fig:restoration}). Figure~\ref{fig:restoration} compares a GPT-image-2.5 gen4 face close-up with the specified forced-restoration output in a vertically stacked native-frame layout, without a separate enlarged crop. In the portrait case series, hair-constrained regeneration produces continuous strands in twelve outputs across six originals and two channel--resolution configurations. These examples demonstrate reconstruction routes, not isolated prompt effects or recovery of a known ground truth.

Reference cleaning also reduces texture carried into a new generation. In a paired example using radial soft clipping for reference preparation, output debris density decreases from 1,842 to 1,020 components per megapixel (45\%). A separate dense-moss comparison gives scale indices of 35.3\% with the untreated reference and 15.8\% with the cleaned reference on the common canvas. Later structure-aware cleaning lowers output band-pass SD from 1.05 to 0.87 in sea and from 2.61 to 1.77 in railing. Sky changes from 0.20 to 0.25, so the effect is region-dependent. These are single-draw comparisons, not average treatment effects; Appendix~\ref{app:protocol} preserves their distinct cleaning protocols and measurements.

\subsection{Selective Suppression with Measured Distortion}
\begin{figure}[t]
\centering
\includegraphics[width=\linewidth]{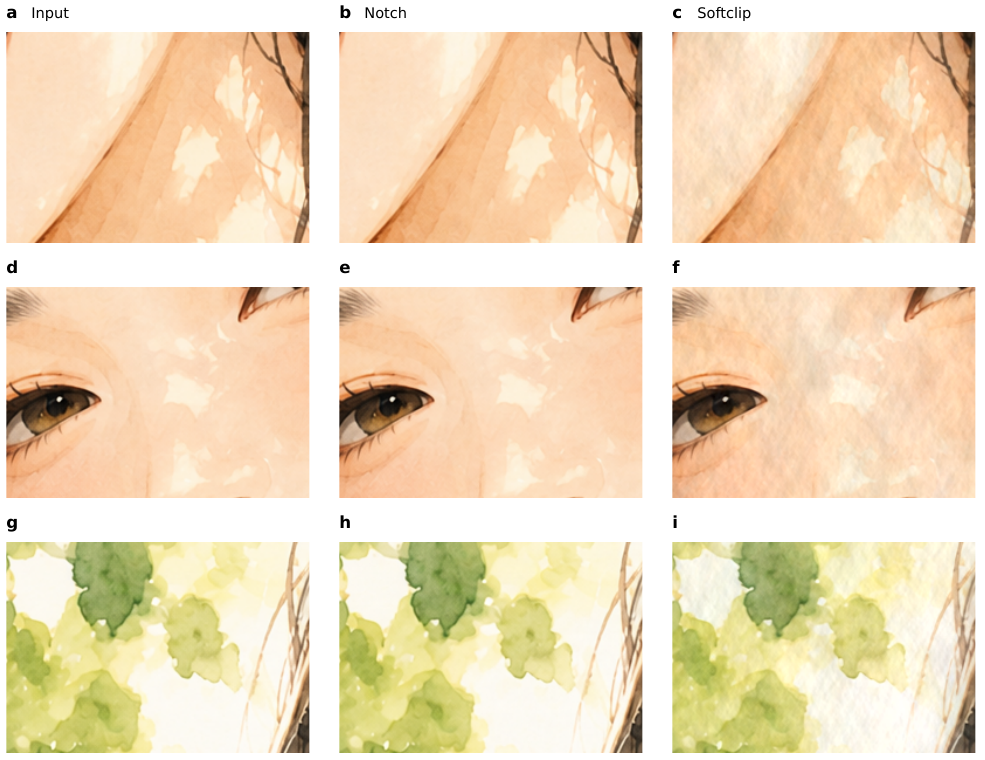}
\caption{Input windows, selective notch outputs and radial soft-clipping outputs for matched skin, eyes and foliage regions.}
\Description{Three columns compare input, notch and soft-clipping outputs across skin, eyes and foliage rows.}
\label{fig:notch}
\end{figure}

\begin{figure}[t]
\centering
\includegraphics[width=\linewidth]{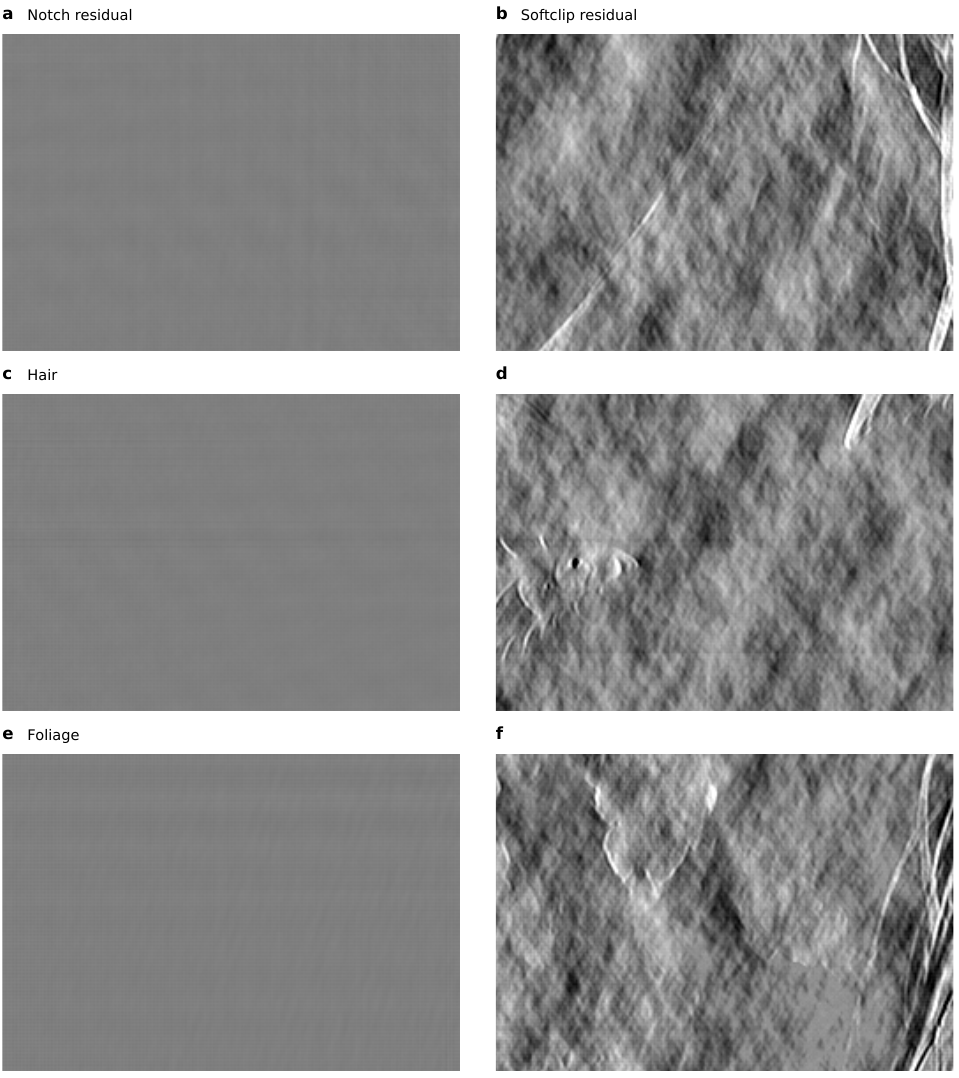}
\caption{Amplified signed residuals for selective notching and radial soft clipping on the matched regions. Gray denotes zero change; residuals are shown with six-fold amplification.}
\Description{Two columns show amplified signed residuals for notch and soft-clipping treatments across skin, eyes and foliage rows.}
\label{fig:notch-residuals}
\end{figure}

Selective notching removes isolated lattice peaks while avoiding the broad tonal changes caused by radial-baseline soft clipping (Figures~\ref{fig:notch} and~\ref{fig:notch-residuals}). On the garden-tilt example, the selected pre-feather mask covers 0.11\% of frequency bins; feathering extends attenuation beyond those bins. A background patch's anomaly decreases from 3.49 to 1.77 with whole-image residual SD 0.18. In the soft-clipping comparison, more than 85\% of removed energy lies within the lowest quarter of the spectral radius.

For local grain, strict masked suppression reduces sky and sea band-pass SD from 0.38 to 0.18 and from 1.21 to 0.71 in one degraded example. The mask protects hair and facial structure. Across the six initial portrait candidates, all detect a lattice and two additionally receive masked suppression. All six pass the distortion checks, with whole-image residual SD 0.21--0.49 and high-frequency retention 98.9--99.8\%. Six additional hair-constrained candidates pass after notching, with residual SD 0.25--0.44. Across fourteen notch-only executions, residual SD spans 0.08--0.44; the range includes repeated use of one candidate rather than fourteen independent images.

\subsection{Routing Across Scene Types}
The restoration route depends on whether the artifact is separable from image content. Smooth or weakly textured regions with isolated spectral peaks can follow selective notching and distortion checks, whereas dense foliage, hair, and other content-entangled regions require masked suppression, cleaned-reference regeneration, or human review. This routing prevents a lower artifact score from being treated as sufficient evidence when filtering may erase legitimate structure.

The eight-scene comparison on the gpt-image-2.5 route provides a supplementary check of this principle. The restoration path reduced six of eight gen4 endpoints to the none grade; moss and wisteria decreased from 25.3\% to 11.1\% and from 41.1\% to 12.1\%, respectively, but remained structured, whereas ice cave decreased from 20.0\% to 0.0\%. These single-chain observations show scene-dependent reduction rather than a universal restoration rate. In two targeted hair/background cases, direct regeneration without reference cleaning failed visual review; cleaning removed the visible defects, while one case introduced a colour shift. Thus, scale-index reduction, visual acceptance, and pixel-aligned recovery remain separate criteria.

\section{Characterization Results}
\label{sec:results}
Table~\ref{tab:evidence-matrix} summarizes the principal evidence using separate rows for spectral anomaly, scale-index coverage, and aligned distortion. Post-treatment interpretation remains in the surrounding text.

\begin{table*}[t]
\caption{Representative measurements and treatments for artifact characterization. Values use the stated measurement convention.}
\label{tab:evidence-matrix}
\centering\small
\begin{tabular}{p{0.20\textwidth}p{0.25\textwidth}p{0.20\textwidth}p{0.25\textwidth}}
\toprule
Evidence case & Observed signature & Measurement & Treatment \\
\midrule
Photographic controls & Low spectral variation & Median anomaly \textbf{1.73} & None \\
Generated-source controls & Elevated spectral anomaly & Medians \textbf{4.13--4.61} & Selective notching \\
Same-scene moss chain & Increasing granular coverage & 0.9\% $\rightarrow$ 23.7\% & Reference cleaning and regeneration \\
Same-scene face chain & No qualifying scale tiles & \textbf{0.0\%} $\rightarrow$ \textbf{0.0\%} & Hair-protected regeneration and notching \\
Portrait restoration & Content-entangled hair structure & SD 0.21--0.49; retention 98.9--99.8\% & Masked suppression and hair-constrained regeneration \\
Image 2.5 moss/wisteria & Structured scale texture & H: 25.3\% $\rightarrow$ 11.1\%; W: 41.1\% $\rightarrow$ 12.1\% & Cleaned-reference regeneration and final notching \\
Image 2.5 ice cave & Structured scale texture & K: 20.0\% $\rightarrow$ \textbf{0.0\%} & Cleaned-reference regeneration and final notching \\
\bottomrule
\end{tabular}
\end{table*}

\subsection{Lattice Signatures Depend on the Access Configuration}
The patch-based survey gives spectral-anomaly medians of 1.73 for photographs and 1.95 for web references, compared with 4.61, 4.17, and 4.13 for the three generation sources. These values are summarized in Table~\ref{tab:baseline}. One photograph reaches 3.12 because of JPEG blocking and repeated structures, so spectral anomaly is not specific to generated images. Four fixed-prompt, fixed-reference draws yield $\max A=4.41,4.49,4.25,4.61$, with a coefficient of variation of 3.4\%.

The channel survey contains 69 images. Channel B is lattice-positive in 43/43 outputs across the tested scenes and resolutions. Channel A is negative in 20/20 outputs at $1280\times720$ and positive in 6/6 at $1536\times1024$. Thus, a vendor name alone does not predict the observed artifact. These counts establish configuration dependence in the sampled conditions, not invariance to every possible prompt or image.

Banana100 provides a complementary observation. Five families exhibit persistent or late-emerging characteristic periods, while four show strong positive step correlations in autocorrelation strength. These sets overlap but are not identical (Appendix~\ref{app:external}). For example, Qwen maintains a six-pixel horizontal period at all ten steps without increasing strength. The GPT sequences remain weakly periodic, unlike some in-house outputs bearing a related vendor label. Different access paths and resolutions prevent a direct vendor ranking. In Flux.2 Max, changed wall texture additionally contributes to the spectral anomaly; periodicity and content drift coexist.

\subsection{Granular Texture Tracks Scene Content and Repetition}
The native-resolution scale-tile percentages for the five same-scene chains are given in Table~\ref{tab:chains}. Face and beach remain at zero qualifying scale tiles in these samples; this indicates that the scale-index probe did not detect this specific texture, not that face restoration is unnecessary. Hair and facial structure are evaluated separately in the portrait cases because they are content-entangled and identity-bearing.

The cross-version comparison shows that the phenomenon persists on the gpt-image-2.5 route but is redistributed across scenes. At gen4, H, W and K reached 25.3\%, 41.1\% and 20.0\%, respectively; F and T were 3.7\% and 2.1\%, while C, G and S were 0.0\%. These single-chain observations do not support a model ranking. The area-CV criterion also distinguishes repeated motifs from convergent equal-sized tiles, and coverage should not be read as visual conspicuity: in the H chain, Image 2.5 had lower relative band contrast despite a comparable unit diameter.

The scene-change rainforest sequence returns to 1.1--2.1\% after a 7.9\% intermediate output; its glacier counterpart remains at 0.0--0.5\%. These observations motivate avoiding unnecessary same-scene recursion. However, scene-change and same-scene sequences also differ in channel, and several same-scene chains mix resolutions. They do not isolate a causal effect of changing the scene.

Granular texture also varies across independent initial draws. Four moss-scene draws span 0.0--16.8\% on the common canvas, whereas four rainforest draws span 9.5--12.6\%. The low-scoring moss draw contains broad leaves instead of dense fine vines. This association supports composition-sensitive quality control, without implying that every dense scene must develop granular artifacts.

\subsection{Prompt Constraints Do Not Provide a Consistent Control}
Eight same-reference pairs compare an unchanged editing prompt with the same prompt plus a foliage-texture constraint. The constraint increases the scale index in five pairs and decreases it in three. A subsequent decomposition tests plant nouns, positive texture properties, and negations separately against the same eight controls. Mean paired differences are $+3.3$, $-1.3$, and $-0.3$ percentage points; two-sided sign-test $p$-values are 0.29, 0.73, and 0.73.

Repeated calls for eleven prompt--reference conditions have a median observed range of 8.4 percentage points. This range contextualizes sampling variability; it is not an equivalence margin or confidence interval. The paired evidence does not establish a consistent benefit for these constraints, nor does it prove that all texture-related prompts are ineffective. Full paired values and the retry protocol appear in Appendix~\ref{app:prompt}.

\section{Discussion and Conclusion}
Diagnosis-guided restoration turns the distinction between lattice and granular artifacts into a practical editing decision. Isolated peaks permit selective filtering with aligned distortion checks; content-entangled texture calls for cleaned-reference regeneration and visual review. Separating these routes connects artifact measurements to restoration choices instead of treating a lower spectral score as the goal.

Configuration-dependent periods are compatible with sampling, decoding, or delivery-path effects, while texture trajectories are compatible with repeated application of a scene-dependent generative prior. A descriptive recurrence, $R_n=I+\alpha R_{n-1}$, expresses new injection and reference carry-over, with fixed point $I/(1-\alpha)$ for constant $I$ and $0\leq\alpha<1$. These parameters are not fitted and do not identify an internal mechanism. The actionable implication is to preserve approved references and avoid unnecessary output-to-input chains; a star-shaped workflow removes direct dependence on the previous output.

\paragraph{Limitations.}
The study-specific thresholds require broader calibration, and mixed-channel, mixed-canvas chains do not isolate causal editing effects. Single-draw regeneration comparisons demonstrate restoration routes rather than population-average gains; regeneration can alter details, and the star-shaped workflow's quality advantage remains unbenchmarked. Matched-channel studies and independently annotated images would establish treatment success rates beyond these cases.

The workflow restores images through three linked decisions: diagnose the artifact, choose a compatible treatment, and verify the candidate. The examples show cleaner reconstructed texture, while aligned filtering achieves small measured residuals across the evaluated executions. Together, these routes provide an explicit path from detecting digital ripple to producing and reviewing a cleaner image.

\paragraph{Disclosure and provenance.}
Channel B is operated by the authors' institution. The experiments concern artifact measurements rather than overall channel rankings. Generated outputs remain identified as AI-generated; filtering is intended for quality control on the authors' assets, not concealment of provenance. Third-party material retains its original attribution and license.

\bibliographystyle{ACM-Reference-Format}
\bibliography{sigplan_refs}

\ifincludeappendix
\clearpage
\appendix
\raggedbottom
\section{Measurement and Treatment Details}
\label{app:measurement}
\subsection{Signal Definitions and Calibration}
All lightness measurements use CIELAB $L^*$ on a 0--100 scale. The nominal 3--8-pixel band-pass is implemented as the difference between Gaussian smooths with standard deviations 1 and 3 pixels. Unless stated otherwise, residual images use a gain of six and middle gray to indicate zero change. These visualizations emphasize weak signals and must not be interpreted as their unamplified appearance.

Spectral-anomaly surveys use several 384-pixel flat patches per image and aggregate their peak scores by the median. The two-chain illustration instead uses a single 512-pixel patch per image. Its values are not pooled with the 384-pixel survey. The autocorrelation probe uses $H=L-G_4L$, a windowed power spectrum, and normalization by its zero-displacement value. It searches outside displacement radius four; opposite vectors represent the same period. The unbandpassed granular-texture check uses a broader $\sigma=8$ illumination-removal step.

The lattice implementation reflection-pads the lightness image by 64 pixels. Candidate components exceed the $21\times21$ local spectral median by 1.2, exclude radius below 24, and contain at most 80 bins. A positive diagnosis requires at least two retained components and a largest excess of at least 2.5; the original channel tables also report a conservative 3.5 threshold. All four photographic controls are measured by this whole-image probe and score 1.39--1.89, independently of flat-patch eligibility. Notching uses Equation~\ref{eq:notch} with 1.5-bin feathering and leaves chroma unchanged by default.

\subsection{Spatial Probes}
Flat-window selection requires a structure-permission mask mean of at least 0.75 and minimum of at least 0.25. Among eligible 96-pixel windows, the six with greatest band-pass energy are examined. A granule-positive window has band-pass SD at least 0.35, excess kurtosis at most 8, medium-blob coverage at least 19\%, and anisotropy at most 0.35. Excess kurtosis is $\langle z^4\rangle-3$, where $z$ is the standardized within-window band-pass response. Grading first assigns pervasive when at least three windows and at least 60\% of examined windows are positive. Otherwise, at least two positive windows yields flat, one yields suspected, and zero yields none. Calibration used fourteen annotated windows, including four positive and ten negative examples.

The whole-frame scale index uses 128-pixel tiles and a 64-pixel stride. Positive and negative band-pass lobes are thresholded separately at one within-tile SD. Connected components of area 12--300 pixels supply blob statistics. A tile qualifies when coverage is at least 19\%, band-pass SD at least 0.30, area CV at most 0.65, mean circularity at least 0.55, and anisotropy at most 0.35. Circularity is $4\pi a/p^2$, clipped at one; the implementation estimates perimeter from boundary pixels. Fewer than five eligible blobs fails the area-variation condition. Anisotropy is $(\lambda_1-\lambda_2)/(\lambda_1+\lambda_2)$ from the integrated band-pass structure tensor.

The scale index is the percentage of qualifying tiles. Scores below 2\% are graded none, scores in $[2\%,6\%)$ suspected, and scores of at least 6\% structured. These thresholds serve within-study triage rather than a validated universal classifier. Missing eligible windows, insufficient image size, and a negative score are distinguished in the diagnostic record. Heat maps accompany aggregate measurements so that detections can be checked against scene content.

Granular regions show weak first autocorrelation maxima at inconsistent displacements; typical measurements have radii of 9--20 pixels and heights below 0.05. By contrast, the approximately 5.17-pixel equivalent blob diameter obtained after band-pass filtering reflects the instrument's preferred scale. We therefore do not report that diameter as a physical period. Blob-area CV also decreases on the beach chain, despite zero scale tiles, showing why a single feature is insufficient.

\subsection{Filtering, Reference Cleaning, and Verification}
Strict spatial suppression subtracts the diagnostic band-pass weighted by a structure-permission mask. Smooth transitions use edge-strength bounds 0.8--2.5, coherence bounds 0.15--0.45, and texture-density bounds 1.8--4.0. Reference-grade cleaning instead uses bounds 1.5--4.0, 0.35--0.70, and 5--10, followed by directional averaging within $\pm10$ pixels at strength 0.85. Frontal and profile face detections are expanded by 25\% and feathered over ten pixels to protect against spatial cleaning. Missed faces remain an operational risk; the image is not an approved deliverable.

Let $H=L-G_1L$, where $G_1$ is Gaussian smoothing with standard deviation one pixel. High-frequency retention is $100\,\sd(H_{\mathrm{out}})/\sd(H_{\mathrm{in}})$, an SD ratio in percent rather than an energy ratio. Acceptance windows are selected during diagnosis and reused after aligned filtering. Regenerated outputs are excluded from this comparison because scene details can change. A passed check means that the measured distortion falls within the study's limits, not that texture, identity, or semantic fidelity has been independently certified.

\section{Experimental Protocol and Additional Controls}
\label{app:protocol}
\subsection{Baseline Survey and Same-Scene Trajectories}
Table~\ref{tab:baseline} retains the patch-based spectral survey; Table~\ref{tab:chains} gives the exact native-resolution values for the five same-scene chains. Three photographs supply valid 384-pixel flat-patch measurements. Chelsea has no qualifying patch and is excluded from that aggregation, but all four photographs are eligible for the whole-image lattice probe.

\begin{table}[t]
\caption{Spectral-anomaly survey under the same 384-pixel patch convention. Each value summarizes one image before aggregation across a group. Photographs have three valid images out of four; Chelsea has no qualifying patch.}
\label{tab:baseline}
\centering\small
\begin{tabular}{lrrr}
\toprule
Source & $n$ & Median & Range \\
\midrule
Photographs & 3 & 1.73 & 1.70--3.12 \\
Web references & 5 & 1.95 & 1.61--2.36 \\
gpt-image-2 metadata & 5 & \textbf{4.61} & 3.61--4.71 \\
Gemini image preview & 5 & \textbf{4.17} & 4.04--4.54 \\
Channel A & 5 & \textbf{4.13} & 2.91--4.50 \\
\bottomrule
\end{tabular}
\end{table}

\begin{table}[t]
\caption{Native-resolution scale-tile percentages over five same-scene outputs on Channel B. Mixed-resolution trajectories are descriptive; canvas dimensions are specified below. Zero values indicate no qualifying scale tiles under the probe, not an absence of other degradations.}
\label{tab:chains}
\centering\small
\begin{tabular}{lrrrrr}
\toprule
Scene & gen0 & gen1 & gen2 & gen3 & gen4 \\
\midrule
Face & \textbf{0.0} & \textbf{0.0} & \textbf{0.0} & \textbf{0.0} & \textbf{0.0} \\
Moss canyon & 0.9 & 3.1 & 5.2 & 8.3 & 23.7 \\
Rainforest & 9.5 & 7.7 & 19.4 & 18.8 & 16.9 \\
Glacier & 1.5 & 1.8 & \textbf{0.6} & \textbf{1.2} & \textbf{1.5} \\
Beach & \textbf{0.0} & \textbf{0.0} & \textbf{0.0} & \textbf{0.0} & \textbf{0.0} \\
\bottomrule
\end{tabular}
\end{table}

\subsection{Sample Accounting and Canvas Dimensions}
The 69-image lattice survey combines 43 channel-B images and 26 channel-A images. Channel B contributes 25 same-scene chain images, six constrained portrait regenerations, and twelve prompt or in-loop outputs. Channel A contributes twenty $1280\times720$ outputs and six $1536\times1024$ initial portrait candidates. The 43 channel-B survey images must not be confused with the separate 43 API calls in the prompt-decomposition experiment.

The five channel-B chains use the same approved character sheet for gen0, then the previous generated output for each subsequent step. Face has four $1280\times720$ outputs and one $1672\times941$ output. Moss has one smaller and four larger outputs; rainforest and beach have the same counts. Glacier has five $1672\times941$ outputs. The two scene-change chains each contain five $1280\times720$ channel-A outputs. Thus, a channel-matched, resolution-matched factorial comparison of scene change remains unperformed.

The original two-chain illustration starts from one $1536\times1024$ generated illustration. One chain changes lighting, while the other changes scene and action. The shared gen0 has spectral anomaly 3.028. The first edit increases it to 3.904 and 4.204, respectively. Dominant period magnitudes remain approximately 4--5 pixels, although orientation varies. Six spatially separated flat patches in the lighting endpoint yield period vectors $(4,-1)$, $(-5,0)$, $(1,-4)$, and $(1,4)$. These observations support a spatially widespread signal but do not uniquely identify one oblique lattice.

\subsection{Single-Draw Composition and Restoration Controls}
Replacing dense fine elements with fewer large structures yields scale indices of 5.8\% for moss and 1.6\% for rainforest. Each is one draw with changed composition. The moss result lies below three high-scoring repeats but above the observed 0.0\% low-tail draw. It is therefore incorrect to describe it as below the entire repeat range. Rewording and adding negative constraints each produce 17.4\% in moss, slightly above the observed 16.8\% maximum, rather than inside that range.

For the moss gen4 input, the scale index is 23.7\% at native $1280\times720$. One further unchanged edit gives 35.3\%. Cleaning the reference before regeneration gives 8.0\% natively and 15.8\% after normalization; appending a foliage constraint gives 7.7\% natively and 9.5\% after normalization. The paired prompt study in Appendix~\ref{app:prompt} prevents interpreting the last single draw as a reliable constraint effect.

One cleaning control uses radial soft clipping as reference preparation, while another uses structure-aware cleaning. Its untreated-reference and cleaned-reference outputs have debris densities of 1,842 and 1,020 per megapixel, and spectral anomalies of 4.303 and 4.048. Debris density counts Canny components of area 2--40 pixels and also responds to legitimate detail. It is used only for comparable scene content, not as a general image-quality score.

In the structure-aware cleaning comparison, output band-pass SD is 1.05 versus 0.87 for sea and 2.61 versus 1.77 for railing, without versus with reference cleaning. Sky is 0.20 versus 0.25; lower energy is not a universal outcome. Hair and face readings contain legitimate structure and are not interpreted as artifact quantities. Notching the cleaned-reference regeneration gives residual SD 0.21 and high-frequency retention 99.7\%.

A restoration without access to a clean ancestor increases the sharpness measure from 3.18 to 4.34. The unseen ancestor scores 5.54, but has different composition, so the ratio is descriptive rather than a recovery rate. Regeneration reconstructs plausible detail; it does not recover a known pixel-level ground truth.

Figure~\ref{fig:restoration} shows the cleaned-reference regeneration before final notching. The separate post-regeneration notching measurement above belongs to the structure-aware cleaning comparison.

\subsection{Cross-Version Endpoint Comparison}
\label{app:crossversion}
Table~\ref{tab:crossversion} reports all eight scenes on the common 1280-pixel-long-edge canvas. Each model--scene cell is one chain and one archived repair outcome, not a repeated trial. The Image 2 routes used notch-only processing for C/G/S and regeneration for the other scenes. Image 2.5 outcomes include forced regeneration branches and separately finalized candidates; a none scale grade is not a certificate of colour, identity or lattice fidelity. H/W requested human review after regeneration before separate final processing. The original 69-image lattice survey and fourteen notch-only executions exclude these new trials.

\begin{table*}[t]
\caption{Cross-version scale-tile percentages. Each trajectory lists gen0 through gen4; repair is the archived processed endpoint. All readings use the same canonical canvas convention. One chain per scene and version.}
\label{tab:crossversion}
\centering\small
\begin{tabular}{llrlr}
\toprule
Scene & Image 2: gen0--gen4 & Repair & Image 2.5: gen0--gen4 & Repair \\
\midrule
C Face & 0.0, 0.0, 0.0, 0.0, 0.0 & 0.0 & 0.0, 0.0, 0.0, 0.0, 0.0 & 0.0 \\
S Beach & 0.0, 0.0, 0.0, 0.0, 0.0 & 0.0 & 0.0, 0.0, 0.0, 0.0, 0.0 & 0.0 \\
G Glacier & 0.5, 0.5, 0.0, 0.0, 0.0 & 0.0 & 1.1, 0.0, 0.0, 0.0, 0.0 & 0.0 \\
T Temple steps & 2.1, 4.2, 6.3, 6.8, 8.9 & 0.0 & 5.3, 7.9, 1.6, 2.1, 2.1 & 0.0 \\
F Rainforest & 9.5, 8.9, 28.4, 22.1, 16.3 & 1.6 & 3.2, 3.2, 2.6, 2.6, 3.7 & 0.0 \\
K Ice cave & 8.9, 24.7, 23.7, 21.1, 23.2 & 0.5 & 0.0, 25.8, 21.6, 20.0, 20.0 & 0.0 \\
H Moss gorge & 0.0, 2.1, 7.4, 8.4, 23.7 & 1.1 & 5.8, 12.6, 27.4, 22.1, 25.3 & 11.1 \\
W Wisteria & 5.8, 11.6, 13.7, 12.1, 13.2 & 4.7 & 13.2, 12.6, 17.4, 14.7, 41.1 & 12.1 \\
\bottomrule
\end{tabular}
\end{table*}

\section{External Seven-Family Sequence Analysis}
\label{app:external}
The Banana100 subset uses one photograph of a wall, table, and lamp. Successive edits add fruit. A wall window is selected at the same relative location in each sequence, retaining native pixels. Window dimensions range from $591\times357$ to $737\times431$ for generated outputs, and are $922\times555$ for the original. Comparisons of absolute spectral values across families therefore have different support sizes.

Table~\ref{tab:external} reports dominant periods, their occurrence counts, and the Spearman correlation between autocorrelation strength and step. Period counts describe stability, not a significance test. Flux.2 Dev shows rising strength but no stable dominant period. Nano Banana 2 Fast has characteristic periods without a consistently increasing autocorrelation peak. These distinctions motivate reporting persistence separately from accumulation.

\begin{table*}[t]
\caption{Native-window periodicity in eleven Banana100 sequences. Period vectors are given as absolute horizontal and vertical displacements. Counts are out of ten steps; $\rho$ refers to autocorrelation strength versus step.}
\label{tab:external}
\centering\small
\begin{tabular}{llcrrr}
\toprule
Family & Sequence & Period (pixels) & Count & Peak first $\to$ last & $\rho$ \\
\midrule
GPT & Different chat & $(0,6)$, weak & 6 & $0.07\to0.06$ & $-0.27$ \\
GPT & Same chat & $(0,6)$, weak & 4 & $0.07\to0.11$ & $0.37$ \\
Flux.2 Dev & Different chat & $(3,3)$, weak & 3 & $0.03\to0.12$ & $1.00$ \\
Flux.2 Max & Different chat & $(0,5)$, late & 5 & $0.05\to0.59$ & $1.00$ \\
Grok Imagine & Different chat & $(0,5)$ & 10 & $0.19\to0.65$ & $0.89$ \\
Grok Imagine & Same chat, left & $(0,5)$ & 10 & $0.17\to0.67$ & $0.95$ \\
Grok Imagine & Same chat, right & $(0,5)$ & 9 & $0.17\to0.87$ & $0.99$ \\
Nano Banana 2 Fast & Different chat & $(8,0)$ & 8 & $0.33\to0.33$ & $0.52$ \\
Nano Banana 2 Fast & Same chat & $(8,0)$ & 7 & $0.26\to0.53$ & $-0.16$ \\
Nano Banana Pro & Different chat & $(5,0)$, late & 6 & $0.07\to0.25$ & $0.99$ \\
Qwen 2511 & Different chat & $(6,0)$ & 10 & $0.43\to0.46$ & $-0.26$ \\
\bottomrule
\end{tabular}
\end{table*}

The GPT outputs have peak heights between 0.06 and 0.16 across the sequence, compared with approximately 0.05 for the photograph. Their weak signal should not be generalized to every GPT access configuration. The Flux.2 Max wall becomes cracked plaster, so its rising spectral anomaly partly reflects changed content. Its late five-pixel period supplies additional evidence of periodic structure. The scale index is nearly zero in this largely flat scene, which is why the wall-specific measurements are used instead.

The source is attributed to Tang et al.~\cite{banana100data} under CC BY-NC-SA 4.0, using the recorded September 6, 2026 snapshot. Seven families and eleven sequences do not supply independent replication across starting scenes. Accordingly, no population prevalence estimate or between-vendor quality ranking is inferred.

\section{Paired Prompt Comparisons}
\label{app:prompt}
Four initial draws in moss and four in rainforest each serve as the reference for a control edit and a constrained edit. The control retains the repeat prompt; the treatment appends a foliage constraint. For the first draw of each scene, the control reuses the existing chain's gen1. Decomposition reuses these same controls with three appended passages: plant nouns (P), positive rendering properties (N), and negations (Q). These are repeated comparisons on eight references, not 24 independent reference images.

\begin{table*}[t]
\caption{Scale-tile percentages at $1280\times720$. An asterisk marks a Lanczos-normalized output; native readings are supplied in Table~\ref{tab:native}. Full denotes the complete foliage constraint; P, N, Q denote its noun, property, and negation parts.}
\label{tab:paired}
\centering\small
\begin{tabular}{lrrrrrrr}
\toprule
Reference & gen0 & Control & Full & P & N & Q & Full $-$ control \\
\midrule
Moss 1 & $0.0^*$ & $2.1^*$ & 17.4 & 12.1 & 10.5 & $5.3^*$ & $15.3$ \\
Moss 2 & 15.3 & 27.4 & 37.4 & 14.2 & $17.9^*$ & 14.7 & $10.0$ \\
Moss 3 & 16.8 & 11.6 & 25.8 & $18.4^*$ & 17.9 & 22.1 & $14.2$ \\
Moss 4 & 14.7 & 40.0 & 27.4 & 34.7 & $15.8^*$ & $16.3^*$ & $-12.6$ \\
Rainforest 1 & 9.5 & $8.9^*$ & $2.6^*$ & $13.2^*$ & $4.7^*$ & $8.4^*$ & $-6.3$ \\
Rainforest 2 & 10.0 & $4.7^*$ & $8.4^*$ & $20.0^*$ & $7.9^*$ & $6.3^*$ & $3.7$ \\
Rainforest 3 & 11.6 & $11.1^*$ & $8.4^*$ & $18.9^*$ & $11.6^*$ & 26.3 & $-2.7$ \\
Rainforest 4 & 12.6 & $5.8^*$ & $10.5^*$ & $6.3^*$ & $15.3^*$ & $10.0^*$ & $4.7$ \\
\bottomrule
\end{tabular}
\end{table*}

\begin{table}[t]
\caption{Native scale indices for the normalized entries in Table~\ref{tab:paired}. Dashes indicate that the common-canvas value is already native. Moss 1 gen0 is 0.9\% natively.}
\label{tab:native}
\centering\small
\begin{tabular}{lrrrrr}
\toprule
Reference & Control & Full & P & N & Q \\
\midrule
Moss 1 & 3.1 & -- & -- & -- & 2.5 \\
Moss 2 & -- & -- & -- & 4.6 & -- \\
Moss 3 & -- & -- & 8.9 & -- & -- \\
Moss 4 & -- & -- & -- & 9.2 & 10.2 \\
Rainforest 1 & 7.7 & 2.8 & 10.5 & 6.5 & 7.1 \\
Rainforest 2 & 0.3 & 4.3 & 12.0 & 0.9 & 1.5 \\
Rainforest 3 & 4.6 & 9.2 & 11.1 & 3.7 & -- \\
Rainforest 4 & 2.5 & 4.3 & 4.0 & 10.8 & 4.0 \\
\bottomrule
\end{tabular}
\end{table}

For P, six differences are positive and two negative; N and Q each have five positive and three negative differences. Two-sided exact binomial sign tests under probability $1/2$ yield $p=0.2891$ and $p=0.7266$, respectively, reported rounded in the main text. These tests assess direction consistency, not equivalence or the magnitude of the mean effect. The three comparisons share controls, and no multiplicity-adjusted significance claim is made.

The decomposition requested 24 outputs but generated 43 through resolution-triggered retries. Eight calls returned the requested $1280\times720$ size; the remaining 35 returned $1672\times941$ or $1672\times940$. Sixteen final comparison entries require normalization. All retry attempts are retained in the evidence archive. Eleven conditions have two or three draws, with a median within-condition range of 8.4 and maximum of 23.6 percentage points. Since retries were triggered by canvas mismatch, these ranges are descriptive variability measures, not a balanced repeatability experiment.

For rejection sampling, one of four initial moss draws falls below 2\%, and none of four rainforest draws falls below 6\%. Substituting $\hat p=1/4$ into $1-(1-p)^N$ gives 0.68 at four draws and 0.90 at eight draws. These are plug-in projections under independent identically distributed draws, not validated success probabilities. Zero successes in four rainforest draws does not imply that a low-scoring outcome is impossible.

\section{Case Results, Implementation, and Evidence Access}
\label{app:pipeline}
\subsection{Six-Original Case Series}
The six watercolour originals depict the same character at $1011\times638$. Each supplies one initial channel-A candidate at $1536\times1024$ and one hair-constrained channel-B candidate at the same resolution. Table~\ref{tab:cases} reports aligned processing results. The two candidate sets are different generations; neither their pixels nor their content should be directly differenced across sets.

\begin{table*}[h!]
\caption{Whole-image distortion after pipeline processing. Initial candidates use channel A; constrained candidates use channel B. Both sets are $1536\times1024$. SD is in $L^*$ units; retention is the high-pass SD ratio.}
\label{tab:cases}
\centering\small
\begin{tabular}{llrrrr}
\toprule
Case & Initial-candidate treatment & SD & Retention (\%) & Constrained SD & Constrained retention (\%) \\
\midrule
Night & Masked reduction + notch & 0.47 & 99.2 & 0.29 & 99.8 \\
Rain & Masked reduction + notch & 0.49 & 98.9 & 0.44 & 99.2 \\
Garden-summer & Notch & 0.26 & 99.7 & 0.28 & 99.7 \\
Winter & Notch & 0.24 & 99.8 & 0.36 & 99.5 \\
Garden-tilt & Notch & 0.21 & 99.8 & 0.25 & 99.8 \\
Sea & Notch & 0.30 & 99.6 & 0.28 & 99.7 \\
\bottomrule
\end{tabular}
\end{table*}

The fourteen notch-only executions comprise four initial candidates, six constrained candidates, three same-scene endpoints (face, glacier, beach), and a separate rule-mode garden-tilt run. They reuse one candidate and include related originals. The minimum residual SD, 0.08, occurs on glacier; face and beach yield 0.13 and 0.12. A separate agent-mode garden-tilt execution additionally applies masked reduction and is not included in the notch-only range.

On matched flat windows, radial soft clipping gives residual SD 1.66--2.78. The four notch-only initial candidates give 0.14--0.25. The two masked-reduction candidates give 0.62--1.13 because they intentionally remove local grain. These are window values, not the whole-image values in Table~\ref{tab:cases}. Keeping both reporting levels prevents conflating local treatment strength with global distortion.

Ribbon-like hair is a rendering change rather than a separable lattice. Across twelve examined hair windows, anisotropy is 0.53--0.69 and spectral energy forms directional lobes rather than an isolated point signature. Strong transverse or isotropic smoothing removes the visible mesh together with strand edges. Hair-constrained regeneration produces continuous strands in twelve outputs across six originals and two channel--resolution configurations. Without matched unconstrained repeats, this demonstrates a viable route, not an isolated prompt effect or a general success rate.

\subsection{Implementation and Release Boundaries}
The implementation separates image I/O, diagnosis, scale indexing, notching, masked suppression, reference cleaning, regeneration, verification, and orchestration. Rule-based routing is the default. An optional language-model decision layer chooses only among permitted actions and cannot change numerical thresholds. Execution records contain observations, allowed actions, decisions, and verification results. Suspected windows are reported for human inspection; regeneration requires explicit permission and a bounded retry count.

The archived report records nineteen passing synthetic and protocol tests. Those tests cover known lattice and granule inputs, clean gradients, the no-flat-window case, distortion limits, and regeneration through a mock client. They are software checks, not independent validation of detector accuracy. One integrated attempt encountered HTTP 503 and stopped without delivery. Archived executions also complete regeneration rounds for moss, rainforest and ice cave. Their records contain source/output hashes and the actions reference cleaning, regeneration, notching, and successful verification. These three qualitative cases are additional to the fourteen notch-only executions; the ice-cave case is not part of the five-chain trajectory experiment. Regeneration changes content and can return a different canvas; the final aligned check measures subsequent filtering damage, not fidelity to the original degraded image.

Measurement and processing modules are supplied in the repository's visual-canon scripts directory. They cover diagnosis, scale indexing, notching, spatial suppression, reference cleaning, regeneration, and verification. The figure process archive retains experiment records, prompts, sidecars, and traces; figure provenance files identify the corresponding source paths. The online-compilation package includes the manuscript figures rather than the complete experimental archive.

All experimental images are existing generated outputs or attributed third-party material. New conference-layout figures rearrange authentic evidence panels and replace annotations with English; they do not synthesize experimental content. Source-panel provenance is retained alongside the figure-generation script. A final submission should supply an appropriately anonymized artifact link if required by the venue.
\fi
\end{document}